\documentclass[twoside]{arxiv}

\usepackage{graphicx}
\usepackage{epstopdf}
\usepackage{booktabs}
\usepackage{enumitem}
\usepackage{verbatim}
\usepackage{dblfloatfix}
\usepackage{algpseudocode}
\usepackage{amsmath,mathtools}
\usepackage{bbding}
\usepackage{balance}
\usepackage{longtable}

\usepackage[most]{tcolorbox}
\newtcolorbox[auto counter]{pabox}[2][]{%
colback=blue!5!white,colframe=blue!75!black,fonttitle=\bfseries,
title=Box~\thetcbcounter: #2,#1}

\begin{document}

\title{Seeing biodiversity: perspectives in machine learning for wildlife conservation}
\shorttitle{Seeing Biodiversity}

\author[1$\dagger$]{Devis Tuia}
\author[1$\dagger$]{Benjamin Kellenberger}
\author[2$\dagger$]{Sara Beery}
\author[3,4,5$\dagger$]{Blair R. Costelloe}
\author[6]{Silvia Zuffi}
\author[7]{Benjamin Risse}
\author[8]{Alexander Mathis}
\author[8]{Mackenzie Weygandt Mathis}
\author[9]{Frank van Langevelde}
\author[10]{Tilo Burghardt}
\author[11,12]{Roland Kays}
\author[13]{Holger Klinck}
\author[3,4]{Martin Wikelski}
\author[3,4,5]{Iain D. Couzin}
\author[13]{Grant van Horn}
\author[3,4,5]{Margaret C. Crofoot}
\author[14]{Charles V. Stewart}
\author[15]{Tanya Berger-Wolf}

\affil[1]{School of Architecture, Civil and Environmental Engineering, Ecole Polytechnique Fédérale de Lausanne (EPFL), Switzerland}
\affil[2]{Department of Computing and Mathematical Sciences, California Institute of Technology (Caltech), United States of America}
\affil[3]{Max Planck Institute of Animal Behavior, Germany}
\affil[4]{Centre for the Advanced Study of Collective Behaviour, University of Konstanz, Germany}
\affil[5]{Department of Biology, University of Konstanz, Germany}
\affil[6]{Institute for Applied Mathematics and Information Technologies, IMATI-CNR, Italy}
\affil[7]{Computer Science Department, University of M\"unster, Germany}
\affil[8]{School of Life Sciences, Ecole Polytechnique Fédérale de Lausanne (EPFL), Switzerland}
\affil[9]{Environmental Sciences Group, Wageningen University, Netherlands}
\affil[10]{Computer Science Department,  University of Bristol, United Kingdom}
\affil[11]{Department of Forestry and Environmental Resources, North Carolina State University, United States of America}
\affil[12]{North Carolina Museum of Natural Sciences, United States of America }
\affil[13]{Cornell Lab of Ornithology, Cornell University, United States of America}
\affil[14]{Department of Computer Science, Rensselaer Polytechnic Institute, United States of America}
\affil[15]{Translational Data Analytics Institute, The Ohio State University, United States of America}
\affil[$\dagger$]{These authors contributed equally}

\leadauthor{Tuia, Kellenberger, Beery, Costelloe}
\maketitle

\begin{abstract}
Data acquisition in animal ecology is rapidly accelerating due to inexpensive and accessible sensors such as smartphones, drones, satellites, audio recorders and bio-logging devices. These new technologies and the data they generate hold great potential for large-scale environmental monitoring and understanding, but are limited by current data processing approaches which are inefficient in how they ingest, digest, and distill data into relevant information. We argue that machine learning, and especially deep learning approaches, can meet this analytic challenge to enhance our understanding, monitoring capacity, and  conservation of wildlife species. Incorporating machine learning into ecological workflows could improve inputs for population and behavior models and eventually lead to integrated hybrid modeling tools, with ecological models acting as constraints for machine learning models and the latter providing data-supported insights. In essence, by combining new machine learning approaches with ecological domain knowledge, animal ecologists can capitalize on the abundance of data generated by modern sensor technologies in order to reliably estimate population abundances, study animal behavior and mitigate human/wildlife conflicts. To succeed, this approach will require close collaboration and cross-disciplinary education between the computer science and animal ecology communities in order to ensure the quality of machine learning approaches and train a new generation of data scientists in ecology and conservation.
\end{abstract}

\section*{Technology to accelerate ecology and conservation research}
Animal diversity is declining at an unprecedented rate~\cite{Ceballos20}. This loss comprises not only genetic, but also ecological and behavioral diversity, and is currently not well understood: out of more than 120,000 species monitored by the IUCN Red List of Threatened Species, up to 17,000 have a `\emph{Data deficient}' status~\cite{IUCNredlist17}. We urgently need tools for rapid assessment of wildlife diversity and population dynamics at large scale and high spatiotemporal resolution, from individual animals to global densities. In this \textit{Perspective} we aim to build bridges across ecology and machine learning to highlight how relevant advances in technology can be leveraged to rise to this urgent challenge in animal conservation.

How are animals currently monitored? Conventionally, management and conservation of animal species are based on data collection carried out by human field workers who count animals, observe their behavior, and/or patrol natural reserves. Such efforts are time-consuming, labor-intensive and expensive~\cite{witmer2005wildlife}. They can also result in biased datasets due to challenges in controlling for observer subjectivity and assuring high inter-observer reliability, and often unavoidable responses of animals to observer presence~\cite{mcevoy2016evaluation, burghardt2012perspectives}. Human presence in the field also poses risks to wildlife~\cite{giese_effects_1996,kondgen_pandemic_2008}, their habitats~\cite{weissensteiner_low-budget_2015}, and humans themselves: as an example, many wildlife and conservation operations are performed from aircraft and plane crashes are the primary cause of mortality for wildlife biologists~\cite{sasse2003job}. Finally, the physical and cognitive limitations of humans unavoidably constrain the number of individual animals that can be observed simultaneously, the temporal resolution and complexity of data that can be collected, and the extent of physical area that can be effectively monitored~\cite{kays2015terrestrial, altmann1974observational}. 

These limitations considerably hamper our understanding of geographic ranges, population densities and community diversity globally, as well as our ability to assess the consequences of their decline. For example, humans conducting counts of seabird colonies~\cite{hodgson2018drones} and bats emerging from cave roosts~\cite{betke2008thermal} tend to significantly underestimate the number of individuals present. Furthermore, population estimates based on extrapolation from a small number of point counts have large uncertainties and can fall victim to nonstationarity: that is, local sampling over a limited time period can fail to capture the spatiotemporal variation in ecological relationships, resulting in erroneous predictions or extrapolations~\cite{rollinson2021working}. Failure to monitor animal populations impedes rapid and effective management actions~\cite{witmer2005wildlife}. For example, insufficient monitoring, due in part to the difficulty and cost of collecting the necessary data, has been identified as a major challenge in evaluating the impact of primate conservation actions~\cite{junker2020severe} and can lead to the continuation of practices that are harmful to endangered species~\cite{sherman2020shifting}. Similarly, poaching prevention requires intensive monitoring of vast protected areas, a major challenge with existing technology. Protected area managers invest heavily in illegal intrusion prevention and the detection of poachers. Despite this, rangers often arrive too late to prevent wildlife crime from occurring~\cite{o2016real}. In short, while a rich tradition of human-based data collection provides the basis for much of our understanding of where species are found, how they live, and why they interact, modern challenges in wildlife ecology and conservation are highlighting the limitations of these methods.

\begin{figure}[!t]
    \centering
    \includegraphics[width=\linewidth]{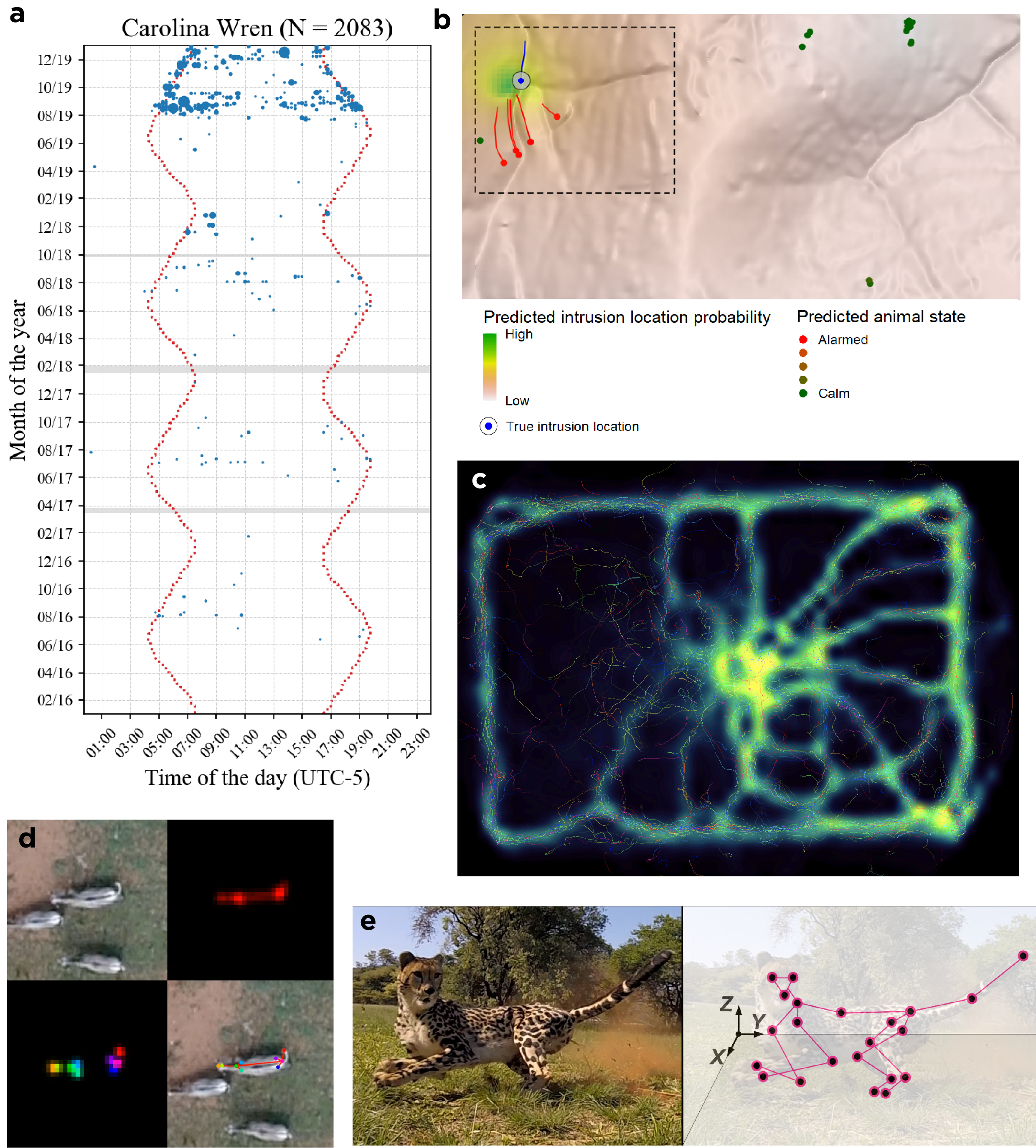}
    \caption{Examples of research acceleration by machine learning-based systems in animal ecology. \textbf{a:} The BirdNET algorithm~\cite{kahl2021birdnet} was used to detect Carolina wren vocalizations in more than 35,000 hours of passive acoustic monitoring data from Ithaca, New York, allowing researchers to document the gradual recovery of the population following a harsh winter season in 2015 (H. Klinck, unpublished). \textbf{b:} Machine-learning algorithms were used to analyze movement of Savannah herbivores fitted with bio-logging devices in order to identify human threats. The method can localize human intruders to within 500m, suggesting `sentinel animals' may be a useful tool in the fight against wildlife poaching~\cite{de2021timely}. \textbf{c:} TRex, a new image-based tracking software, can track the movement and posture of hundreds of individually-recognized animals in real-time. Here the software has been used to visualize the formation of trails in a termite colony~\cite{walter_trex_2021}. \textbf{d, e:} Pose estimation software, such as DeepPoseKit (d)~\cite{graving2019deepposekit} and DeepLabCut (e)~\cite{mathis2018deeplabcut, joska2021acinoset} allows researchers to track the body position of individual animals from video imagery, including drone footage, and estimate 3D postures in the wild. See Acknowledgements for credits and permissions.  
}
\label{fig:sentinel}
\end{figure}

Recent advances in sensor technologies are drastically increasing data collection capacity by reducing costs and expanding coverage relative to conventional methods (see `New sensors expand available data types for animal ecology', below), thereby opening new avenues for ecological studies at scale (Figure~\ref{fig:sentinel})~\cite{lahoz-monfort_comprehensive_2021}. Many previously inaccessible areas of conservation interest can now be studied through the use of high-resolution remote sensing~\cite{Gottschalk05}, and large amounts of data are being collected non-invasively by digital devices such as camera traps~\cite{steenweg2017scaling}, consumer cameras~\cite{hausmann2018social} and acoustic sensors~\cite{sugai_terrestrial_2018}. New on-animal bio-loggers, including miniaturized tracking tags~\cite{wikelski2007going,belyaev2020development} and sensor arrays featuring accelerometers, audiologgers, cameras, and other monitoring devices document the movement and behavior of animals in unprecedented detail~\cite{harel2021locomotor}, enabling researchers to track individuals across hemispheres and over their entire lifetimes at high temporal resolution and thereby revolutionizing the study of animal movement (Figure~\ref{fig:sentinel}c) and migrations.

There is a mismatch between the ever-growing volume of raw measures (videos, images, audio recordings) acquired for ecological studies and our ability to process and analyze this multi-source data to derive conclusive ecological insights rapidly and at scale. Effectively, ecology has entered the age of big data and is increasingly reliant on sensors, methodologies, and computational resources~\cite{farley2018situating}. Central challenges to efficient data analysis are the sheer volume of data generated by modern collection methods and the heterogeneous nature of many ecological datasets, which preclude the use of simple automated analysis techniques~\cite{farley2018situating}. Crowdsourcing platforms like eMammal (\href{https://emammal.si.edu/}{emammal.si.edu}), Agouti (\href{https://www.agouti.eu/}{agouti.eu}) and Zooniverse (\href{https://www.zooniverse.org/}{www.zooniverse.org}) function as collaborative portals to collect data from different projects and provide tools to volunteers to annotate images \emph{e.g.}, with species labels of the individuals therein. Use of such platforms drastically reduces the cost of data processing (\emph{e.g.}~\cite{lasky2021candid} reports a reduction of seventy thousand dollars), but the rapid increase in the volume and velocity of the data collection is making such approaches unsustainable. For example, in August 2021 the platform Agouti hosted 31 million images, of which only 1.5 million were annotated. This is mostly due to the manual nature of the current annotation tool, which requires human review of every image. In other words, methods for automatic cataloging, searching, and converting data into relevant information are urgently needed and have the potential to broaden and enhance animal ecology and wildlife conservation in scale and accuracy, address prevalent challenges, and pave the way forward towards new, integrated research directives.

Machine learning (ML, see glossary in Supplement) deals with learning patterns from data~\cite{hastie01_elements}. Presented with large quantities of inputs (\emph{e.g.}, images) and corresponding expected outcomes, or \emph{labels} (\emph{e.g.}, the species depicted in each image), a supervised ML algorithm learns a mathematical function leading to the correct outcome prediction when confronted with new, unseen inputs. When the expected outcomes are absent, the (this time unsupervised) ML algorithm  will use solely the inputs to extract groups of data points corresponding to typical patterns in the data. ML has emerged as a promising means of connecting the dots between big data and actionable ecological insights~\cite{christin2019applications} and is an increasingly popular approach in ecology~\cite{kwok2019ai,kwok2019deep}, especially due to increased computational power and rapid software advances of the last ten years. A significant share of this success can be attributed to deep learning (DL~\cite{LeCun15}), a family of highly versatile ML models based on artificial neural networks that have shown superior performance across the majority of ML use cases (see Table~\ref{tab:tools} and Supplement). Significant error reduction of ML and DL with respect to traditional generalized regression models has been reported routinely for species richness and diversity estimation~\cite{pichler2020machine,knudby2010predictive}. Likewise, detection and counting pipelines moved from rough rule of thumb extrapolations from visual counts in national parks to ML-based methods with high detection rates. Initially, these methods proposed many false positives which required further human review~\cite{rey2017detecting}, but recent methods have been shown to maintain high detection rates with significantly fewer false positives~\cite{beery2019efficient}. As an example, large mammal detection in the Kuzikus reserve in 2014 was improved significantly by improving the detection methodologies, from a recall rate of 20\%~\cite{rey2017detecting}, to 80\%~\cite{kellenberger2019few} (for a common 75\% precision rate). Finally, studies involving human operators demonstrated that ML enabled massive speedups in complex tasks such as individual and species recognition~\cite{schofield2019chimpanzee, ditria_automating_2020} and large-scale tasks such as animal detection in drone surveys~\cite{kellenberger202121}.
{Recent advances in ML methodology could accelerate and enhance various stages of the traditional ecological research pipeline (see Figure~\ref{fig:ecocycle}), from targeted data acquisition to image retrieval and semi-automated population surveys. As an example, the initiative Wildlife Insights~\cite{ahumada2020wildlife} is now processing millions of camera trap images automatically (17 million in August 2021), providing wildlife conservation scientists and practitioners with the data necessary to study animal abundances, diversity and behavior. Besides pure acceleration, use of ML also massively reduces analysis costs, with reduction factors estimated between 2 and 10~\cite{eikelboom2019improving}.

A growing body of literature promotes the use of ML in various ecological subfields by educating domain experts about ML approaches~\cite{weinstein_computer_2018, valletta_applications_2017, christin2019applications}, their utility in capitalizing on big data~\cite{farley2018situating, peters_harnessing_2014}, and, more recently, their potential for ecological inference (\emph{e.g.}, understanding the processes underlying ecological patterns, rather than only predicting the patterns themselves)~\cite{yu_study_2021, lucas_translucent_2020}. Clearly, there is a growing interest in applying ML approaches to problems in animal ecology and conservation. We believe that the challenging nature of ecological data, compounded by the size of the datasets generated by novel sensors and the ever-increasing complexity of state-of-the-art ML methods, favor a \emph{collaborative approach} that harnesses the expertise of both the ML and animal ecology communities, rather than an application of off-the-shelf ML methodologies to ecological challenges.  Hence, the relation between ecology and ML should not be unidirectional: integrating ecological domain knowledge into ML methods is essential to designing models that are
accurate in the way they describe animal life. As demonstrated by the rising field of hybrid environmental  algorithms (leveraging both DL and bio-physical models~\cite{Reichstein2019,CampsValls21wiley}) and, more broadly, by theory-guided data science~\cite{Karpatne17}, such hybrid models tend to be less data-intensive, avoid incoherent predictions and are generally more interpretable than purely data-driven models. To reach this goal of an integrated science of ecology and ML, both communities need to work together to develop specialized datasets, tools and knowledge. With this objective in mind, we review recent efforts at the interface of the two disciplines, particularly focused on animal ecology and wildlife conservation, present success stories of such symbiosis, and sketch an agenda for the future of the field.

\begin{figure}[!h]
\includegraphics[width = \linewidth]{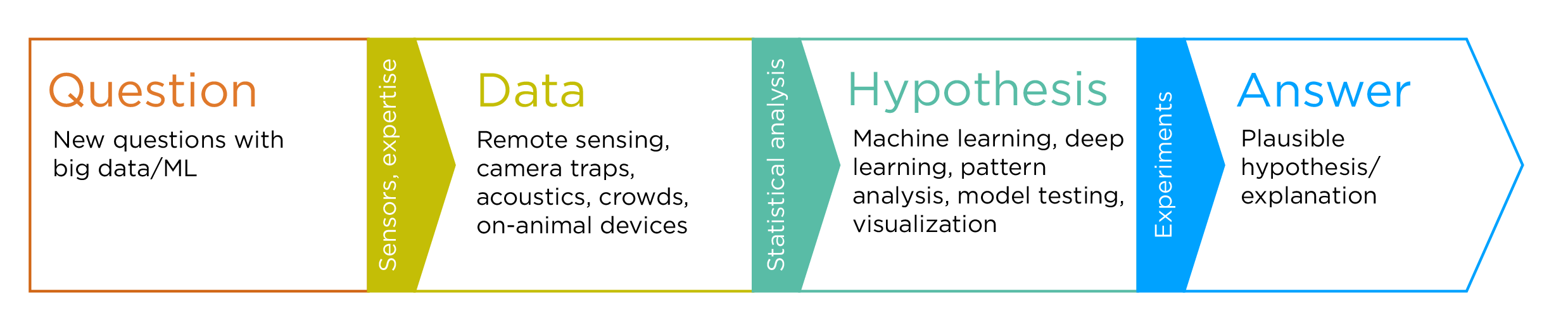}
\caption{Traditional ecological research pipeline (colored text and boxes) and contributions of ML to the different stages discussed in this paper (black text).
}
\label{fig:ecocycle}
\end{figure}

\section*{New sensors expand available data types for animal ecology}
Sensor data provide a variety of perspectives to observe wildlife, monitor populations and understand behavior. They allow the field to scale studies in space, time, and across the taxonomic tree and, thanks to open science projects (Table~\ref{tab:csprojects}), to share data across parks, geographies and the globe~\cite{oliver2020global}. Sensors generate diverse data types, including imagery, soundscapes, and positional data (Figure~\ref{fig:sensors}). They can be mobile or static, and can be deployed to collect information on individuals or species of interest (\emph{e.g.}, bio-loggers, drones), monitor activity in a particular location (\emph{e.g.}, camera traps and acoustic sensors), or document changes in habitats or landscapes over time (satellites, drones). Finally, they can also be opportunistic, as in the case of community science. Below, we discuss the different categories of sensors and the opportunities they open for ML-based wildlife research.

\begin{figure*}
\centering
\includegraphics[width=.9\linewidth]{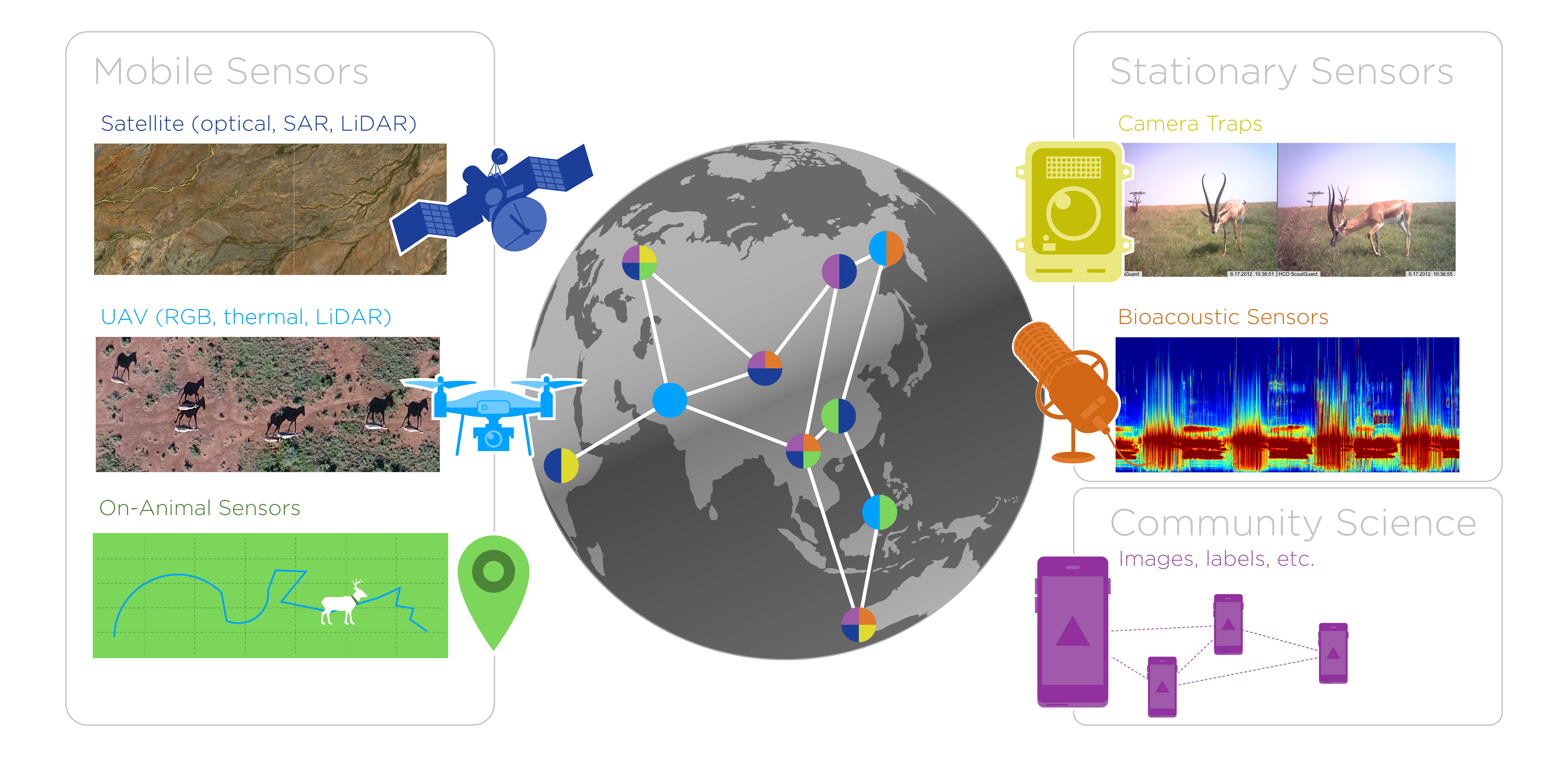}
\caption{A variety of sensors used in animal ecology. Studies frequently combine data from multiple sensors at the same geographic location, or data from multiple locations to achieve deeper ecological insights.}
\label{fig:sensors}
\end{figure*}

\paragraph{Stationary sensors.}
Stationary sensors provide close-range continuous monitoring over long time scales. Sensors can be image-based (\emph{e.g.}, camera traps) or signal-based (\emph{e.g.}, sound recorders). Their high level of temporal resolution allows for detailed analysis, including presence/absence, individual identification, behavior analysis and predator-prey interaction. However, because of their stationary nature, their data is highly spatiotemporally correlated. Based on where and when in the world the sensor is placed, there is a limited number of species that can be captured. Furthermore, many animals are highly habitual and territorial, leading to very strong correlations between data taken days or even weeks apart from a single sensor~\cite{beery2020context}.

\begin{itemize}

\item \emph{Camera traps} are among the most used sensors in recent ML-based animal ecology papers, with more than a million cameras already used to monitor biodiversity worldwide~\cite{steenweg2017scaling}. Camera traps are inexpensive, easy to install, and provide high-resolution image sequences of the species that trigger them, sufficient to specify the animal species, sex, age, health, behavior, and predator-prey interactions. Coupled with population models, camera trap data has also been used to estimate species occurrence, richness, distribution and density~\cite{steenweg2017scaling}. But the popularity of camera traps also creates challenges relative to the quantity of images and the need for manual annotation of the collections: software tools easing the annotation process are appearing (see \emph{e.g.}, AIDE in Table~\ref{tab:tools}) and solutions to filter out empty images or predict species from camera trap images are becoming popular~\cite{norouzzadeh2018automatically,beery2020context,schneider2019past}: ML has proven to be a valuable and robust tool for filtering blank images, and many ecologists have already incorporated open source ML approaches for blank filtering (such as the Microsoft AI4Earth MegaDetector~\cite{beery2019efficient}, see Table~\ref{tab:tools} and Box~\ref{box:md}) into their camera trap workflows. However, problems related to lack of generality across geographies, day/night acquisition or sensors are still major obstacles to production-ready accurate systems~\cite{beery2018recognition}.
The increased scale of available data due to de-siloing efforts from organizations like Wildlife Insights (\href{https://www.wildlifeinsights.org/}{www.wildlifeinsights.org}) and LILA.science (\href{https://www.lila.science/}{www.lila.science}) will help increase ML accuracy and robustness across regions and taxa.

\item\textit{Bioacoustic sensors} are an alternative to image-based systems, using microphones and hydrophones to study vocal animals and their habitats~\cite{sugai2019terrestrial}. Networks of static bioacoustic sensors, used for passive acoustic monitoring (PAM), are increasingly applied to address conservation issues in terrestrial~\cite{wrege2017acoustic}, aquatic~\cite{desjonqueres2020passive}, and marine~\cite{davis2017long} ecosystems. Compared to camera traps, PAM is mostly unaffected by light and weather conditions (some factors like wind still play a role), senses the environment omnidirectionally, and tends to be cost-effective when data needs to be collected at large spatiotemporal scales with high resolution~\cite{wood2019detecting}. While ML has been extensively applied to camera trap images, its application to long-term PAM datasets spanning months to years is still in its infancy and the first DL-based studies are only starting to appear (see Fig~\ref{fig:sentinel}a, \cite{kahl2021birdnet}). Significant challenges remain when utilizing PAM. First and foremost among these challenges is the size of data acquired. Given the often continuous and high-frequency acquisition rates, datasets often exceed the terabyte scale. Handling and analyzing these datasets efficiently requires access to advanced computing infrastructure and solutions. Second, the inherent complexity of soundscapes requires noise-robust algorithms that generalize well and can separate and identify many animal sounds of interest from confounding natural and anthropogenic signals in a wide variety of acoustic environments~\cite{stowell2019automatic}. The third challenge is the lack of large and diverse labeled datasets. As for camera trap images, species- or region-specific characteristics (\emph{e.g.}, regional dialects~\cite{ford2018dialects}) affect algorithm performance. Robust, large-scale datasets have begun to be curated for some animal groups (\emph{e.g.}, \href{https://www.macaulaylibrary.org}{www.macaulaylibrary.org} and \href{https://www.xeno-canto.org}{www.xeno-canto.org} for birds), but for many animal groups as well as relevant biological and non-biological confounding signals, such data is still nonexistent.
\end{itemize}

\paragraph{Remote sensing.}
Collecting data on free ranging wildlife has been restricted traditionally by the limits of manual data collection (\emph{e.g.}, extrapolating transect counts), but have increased greatly through the automation of remote sensing~\cite{rey2017detecting}. Using remote sensing, \emph{i.e.}, sensors mounted on moving platforms such as drones, aircraft, or satellites -- or attached to the animals themselves -- allows us to break these limitations, since the sensors become mobile, and as such enables monitoring of large areas and movement tracking over time.

\begin{itemize}
\item \emph{On-animal sensors} are the most common remote sensing devices deployed in animal ecology~\cite{kays2015terrestrial}. They are primarily used to acquire movement trajectories (\emph{i.e.}, GPS data) of animals, which can then be classified into activity types that relate to the behavior of individuals or social groups~\cite{hughey2018challenges,kays2015terrestrial}. Secondary sensors, such as microphones, video cameras, heart rate monitors and accelerometers, allow researchers to capture environmental, physiological, and behavioral data concurrently with movement data~\cite{williams2020optimizing}. However, power supply and data storage and transmission limitations of bio-logging devices are driving efforts to optimize sampling protocols or pre-process data in order to conserve these resources and prolong the life of the devices. For example, on-board processing solutions can use data from low-cost sensors to identify behaviors of interest and engage resource-intensive sensors only when these behaviors are being performed~\cite{korpela2020machine}. Other on-board algorithms classify raw data into behavioral states to reduce the volume of data to be transmitted~\cite{yu_evaluation_2021}. Various supervised ML methods have shown their potential in automating behavior analysis from accelerometer data~\cite{browning2018predicting,liu2019deep}, identifying behavioral state from trajectories~\cite{wang_machine_2019} and predicting animal movement~\cite{wijeyakulasuriya_machine_nodate}.

\item \emph{Unmanned aerial vehicles (UAVs)}  or drones for low-altitude image-based approaches, have been highlighted as a promising technology for animal conservation~\cite{Linchant15,hodgson2016precision}. Recent studies have shown the promise of UAVs and deep learning for posture tracking~\cite{mathis2018deeplabcut, graving2019deepposekit,mathis2020primer}, semi-automatic detection of large mammals \cite{kellenberger2018detecting,eikelboom2019improving}, birds~\cite{Kel20birds} and, in low altitude flight, even identification of individuals~\cite{andrew2019aerial}. Drones are agile platforms that can be deployed rapidly -- theoretically on demand -- and with limited cost. Thus, they are ideal for local population monitoring. Lower altitude flights in particular can provide oblique view points that partially mitigate occlusion by vegetation. The reduced costs and operation risks of UAVs further make them an increasingly viable alternative to low-flying manned aircraft. In large-scale acquisition scenarios where airplanes are still required, ML concepts can likewise be applied if airplanes are equipped with imaging sensors accordingly. 

Common multi-rotor UAV models are built using inexpensive hardware and consumer-level cameras, and only require a trained pilot with flight permissions to perform the survey. To remove the need for a trained pilot, fully autonomous UAV platforms are also being investigated~\cite{andrew2019aerial}.
However, multi-rotor drone-based surveys remain limited in the spatial footprint that can be covered, mostly because of battery limitations (which become even more stringent in cold climates like Antarctica) and local legislation. Combustion-driven fixed wing UAVs flying at high altitudes can overcome some of these limitations, but are significantly more costly and preclude close approaches for visual measurements on animals. Finally, using drones also has a risk of modifying the behavior of the animals. A recent study~\cite{schroeder2020experimental} showed that flying at lower altitudes (\emph{e.g.}, lower than 150 m) can have a significant impact on group and individual behavior of mammals, although the severity of wildlife disturbance from drone deployments will depend heavily on the focal species, the equipment used, and characteristics of the drone flight (such as approach speed and altitude)~\cite{bennitt2019terrestrial} -- this is a rapidly changing field and advances that will limit noise are likely to come. More research to quantify and qualify such impacts in different ecosystems is timely and urgent, to avoid both biased conclusions and increased levels of animal stress.

\item \emph{Satellite data} is used to widen the spatial footprint and reduce invasive impact on behavior. 
Public programs such as Landsat and Sentinel provide free and open imagery at medium resolution (between 10 and 30 m per pixel), which, though usually not sufficient for direct wildlife observations, can be useful for studying their habitats~\cite{deneu2019evaluation,knudby2010predictive}. Meanwhile, commercial very high resolution (less than one meter per pixel) imagery is narrowing the gap between UAV acquisitions and large-scale footprinting with satellites. Remote sensing has a long tradition of application of ML algorithms. Thanks to the raster nature of the data, remote sensing has fully adopted the new DL methods~\cite{zhu17deep}, which are nowadays entering most fields of application that exploit satellite data~\cite{CampsValls21wiley}. In animal ecology, studies focused on large animals such as whales~\cite{guirado2019whale} or elephants~\cite{duporge2020using} attempt direct detection of the animals on very high resolution images, increasingly with DL. When focusing on smaller-bodied species, studies resort to aerial surveys to increase resolution in order to directly visualize the animals or focus on the detection of proxies instead of the detection of the animal itself (\emph{e.g.}, the detection of penguin droppings to locate colonies~\cite{fretwell2020discovery}). More research is currently required to really harness the power of remote sensing data, which lies, besides the large footprint and image resolution, in the availability of image bands beyond the visible spectrum. These extra bands are highly appreciated in plant ecology~\cite{brodrick2019uncovering} and multi- and hyperspectral DL approaches~\cite{audebert2019deep} are yet to be deployed in animal ecology, where they could help advancing the characterization of habitats. 
\end{itemize}

\subsection*{Community science for crowd-sourcing data}

\justify An alternative to traditional sensor networks (static or remote) is to engage community members as wildlife data collectors and processors~\cite{mckinley2017citizen,waldchen2018machine}. In this case, community participants (often volunteers) work to \emph{collect the data} and/or \emph{create the labels} necessary to train ML models. Models trained this way can then be used to bring image recognition tasks to larger scale and complexity, from filtering out images without animals in camera trap sequences to identifying species or even individuals. Several annotation projects based on community science have appeared recently (Table~\ref{tab:csprojects}). For simpler tasks like animal detection, community science effort can be open to the public, while for more complex ones such as identifying bird species with subtle appearance differences (``fine-grained classification'', also see the glossary), communities of experts are needed to provide accurate labels. \\

\begin{figure}[b]
\begin{center}
\includegraphics[width=\linewidth]{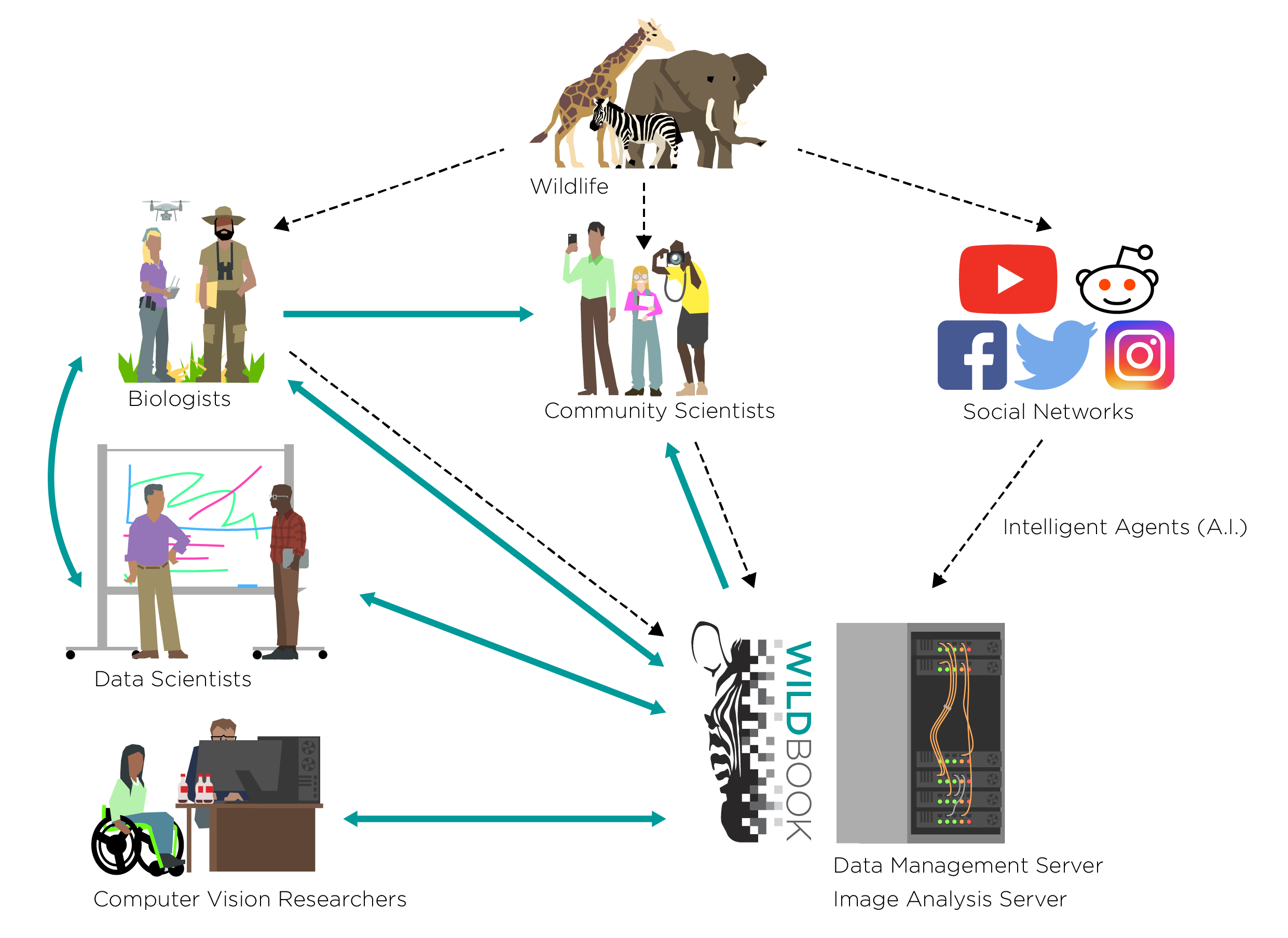}
\end{center}
\captionof{figure}{The Wildbook Ecosystem. Wildbook allows scientists and wildlife managers to leverage the power of communities and machine learning to monitor wildlife populations. Images of target species are collected via research projects, community science events (\emph{e.g.} the Great Grevy's Rally; see text), or by scraping social media platforms using Wildbook AI tools (dotted arrows). Wildbook software uses computer vision technology to process the images, yielding species and individual identities for the photographed animals. This information is stored in databases on Wildbook data management servers. The data and biological insights generated by Wildbook facilitates exchange of expertise between biologists, data scientists, and stakeholder communities around the world (solid arrows).}
\label{fig:Wildbook}
\end{figure}

\begin{pabox}[label={box2},nameref={Hardware}]{Wildbook: successes at the interface between community science and deep learning.}\label{txtboxWB}
Wildbook, a project of the non-profit Wild Me, is an open source software platform that blends structured wildlife research with artificial intelligence, community science, and computer vision to speed population analysis and develop new insights to help conservation (Figure~\ref{fig:Wildbook}). Wildbook supports collaborative mark-recapture, molecular ecology, and social ecology studies, especially where community science and artificial intelligence can help scale up projects. The image analysis of Wildbook can start with images from any source –- scientists, camera traps, drones, community scientists, or social media –- and use ML and computer vision to detect multiple animals in the images \cite{parhamAnimalDetectionPipeline2018} to not only classify their species, but identify individual animals applying a suite of different algorithms \cite{arzoumanian2005astronomical,Weideman_2020_WACV}. Wildbook provides a technical solution for wildlife research and management projects for non-invasive individual animal tracking, population censusing, behavioral and social population studies, community engagement in science, and building a collaborative research network for global species. There are currently Wildbooks for over 50 species, from sea dragons to zebras, spanning the entire planet. More than 80 scientific publications have been enabled by Wildbook. Wildbook data has become the basis for the IUCN Red List global population numbers for several species, and supported the change in conservation status for whale sharks from ``vulnerable'' to ``endangered''. Wildbook's technology also enabled the Great Grevy's Rally, the first ever full species census for the endangered Grevy's zebra in Kenya, using photographs captured by the public. Hosted for the first time in January 2016, it has become a regular event, held every other year. Hundreds of people, from school children and park rangers, to Nairobi families and international tourists, embark on a mission to photograph Grevy's zebras across its range in Kenya, capturing approximately 50,000 images over the two-day event. With the ability to identify individual animals in those images, Wildbook can enable an accurate population census and track population trends over time. The Great Grevy's Rally has become the foundation of the Kenya Wildlife Service's Grevy's zebra endangered species management policy and generates the official IUCN Red List population numbers for the species. Wildbook's AI enables science, conservation and global public engagement by bringing communities together and working in partnership to provide solutions that people trust.
\end{pabox}

A particularly interesting case is Wildbook (see the text box on Page~\pageref{txtboxWB} and Table~\ref{tab:tools}), which routinely screens videos from social media platforms with computer vision models to identify individuals; community members (in this case video posters) are then queried in case of missing or uncertain information. Recent research shows that ML models trained on community data can perform as well as annotators~\cite{torney2019comparison}. However, it is prudent to note that the viability of community science services may be limited depending on the task and that oftentimes substantial efforts are required to verify volunteer-annotated data. This is due to annotator errors, including misdetected or mislabeled animals due to annotator fatigue or insufficient knowledge about the annotation task, as well as systematic errors from adversarial annotators.}

Another form of community science is the usage of \emph{images acquired by volunteers}: in this case, volunteers replace camera traps or UAVs and provide the raw data used to train the ML model. Although this approach sacrifices control over image acquisitions and is likewise prone to inducing significant noise to datasets, for example through low-quality imagery, it provides a substantial increase in the number of images and the chances of photographing species or single individuals in different regions, poses and viewing angles. Community science efforts also increase public engagement in science and conservation. The Great Grevy's Rally, a community science-based wildlife census effort occurring every two years in Kenya~\cite{parham2017animal}, is a successful demonstration of the power of community science-based wildlife monitoring.

\begin{figure*}
\centering
\includegraphics[width=.7\linewidth]{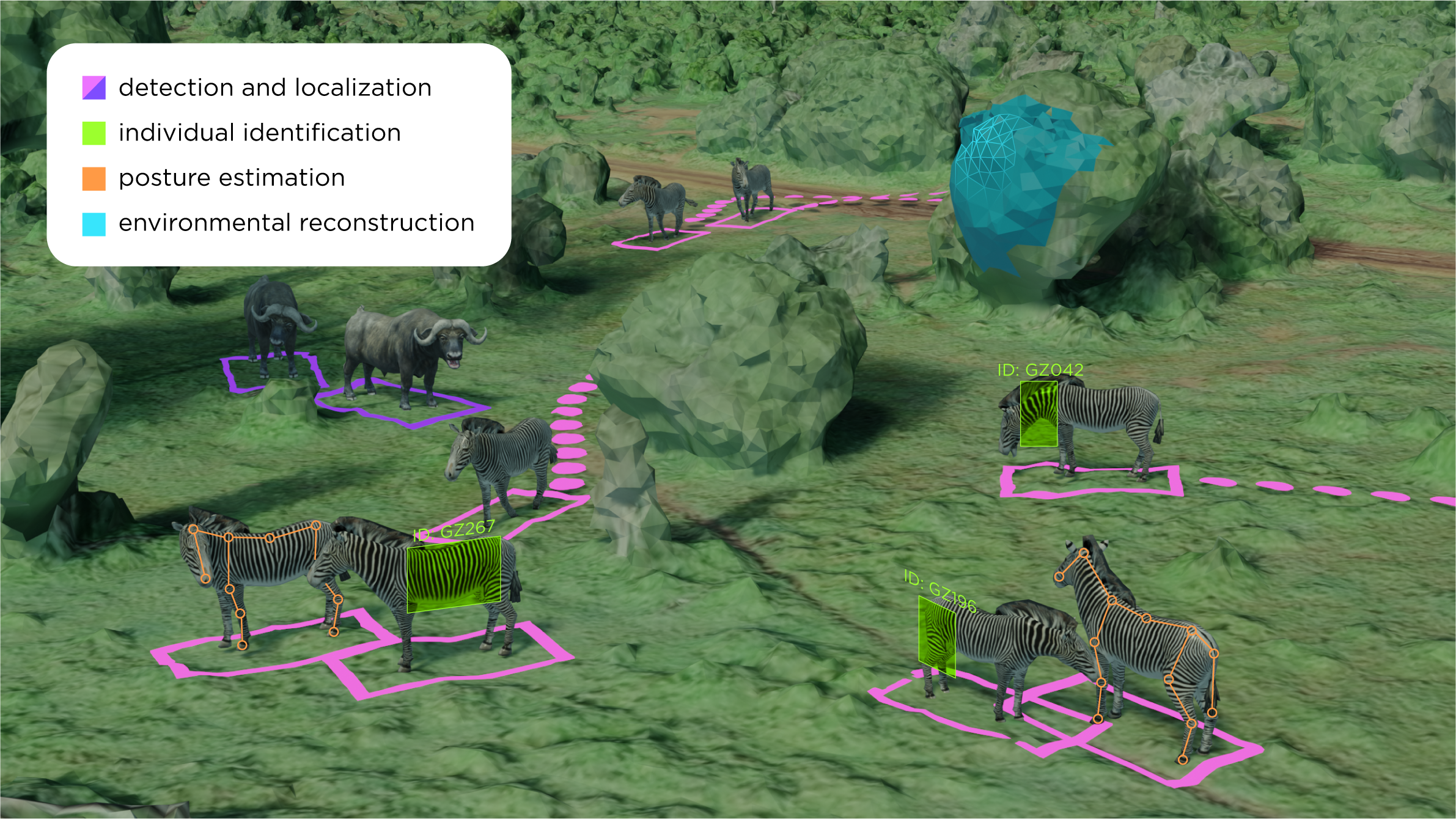}
\caption{\textbf{Setting a common vocabulary: }ecology tasks vs corresponding ones in computer vision. Imagery can be used to capture a range of behavioral and ecological data, which can be processed into usable information with ML tools. Aerial imagery (from drones, or satellites for large species) can be used to localize animals and track their movements over time (pink and purple), and model the 3D structure of landscapes using photogrammetry (blue). Posture estimation tools allow researchers to estimate animal postures (orange), which can then be used to infer behaviors using clustering algorithms. Finally, computer vision techniques allow for the identification and re-identification of known individuals across encounters (green).}
\label{fig:tasks}
\end{figure*}

\section*{Machine learning to scale-up and automate animal ecology and conservation research}

The sensor data described in the previous section has the potential to unlock ecological understanding on a scale difficult to imagine in the recent past. But to do so, it must be interpreted and converted to usable information for ecological research. For example, such conversion can take the form of abundance mapping, individual animal re-identification, herd tracking, or digital reconstruction (three dimensional, phenotypical) of the environment the animals live in. The measures yielded by this conversion, reviewed in this section, are also sometimes referred to as animal biometrics~\cite{kuhl2013animal}. Interestingly, the tasks involved in the different approaches show similarities with traditional tasks in ML and computer vision (\emph{e.g.}, detection, localization, identification, pose estimation), for which we provide a matching example in animal ecology in Figure~\ref{fig:tasks}.

\subsection*{Wildlife detection and species-level classification}

\justify Conservation efforts of endangered species require knowledge on how many individuals of the species in question are present in a study area. Such estimations are conventionally realized with statistical occurrence models that are informed by sample-based species observations. It is these observations where imaging sensors (camera traps, UAVs, \emph{etc.}), paired with  ML models that detect and count individuals in the imagery, can provide the most significant input. Early works used classical supervised ML algorithms (algorithms needing a set of human-labeled annotations to learn, see Supplementary~Table~2): these algorithms were used to make the connection between  a set of characteristics of interest extracted from the image (visual descriptors, \emph{e.g.}, color histograms, spectral indices, \emph{etc.}, also see the glossary)  and the annotation itself (presence of an animal, species, etc.)~\cite{yu2013automated,rey2017detecting}. Particularly in camera trap imagery, foreground (animal) segmentation is occasionally performed as a pre-processing step to discard image parts that are potentially confusing for a classifier. These approaches, albeit good in performance, suffer from two limitations: first, the visual descriptors need to be specifically tailored to the problem and dataset at hand. This not only requires significant engineering efforts, but also bears the risk of the model becoming too specific to the particular dataset and external conditions (\emph{e.g.,} camera type, background foliage amount and movement type) at hand. Secondly, computational restrictions in these models limit the number of training examples, which is likely to have detrimental effects on variations in data (temporal, seasonal, \emph{etc.}), thus reducing the generalization capabilities to new sensor deployments or regions. For these reasons, detecting and classifying animal species with DL for the purpose of population estimates is becoming increasingly popular for images~\cite{norouzzadeh2018automatically,beery2020context}, acoustic spectrograms~\cite{mac2018bat}, and videos~\cite{schindler2021identification}. Models performing accurately and robustly on specific classes (\emph{e.g.} the MegaDetector or AIDE; see Table~\ref{tab:tools}) are now being used routinely and integrated within open systems supporting ecologists performing both labeling and detection, respectively counting of their image databases.
Issues related to dependence of the models performance to specific training locations are still at the core of current developments~\cite{beery2020context}, an issue known in ML as ``domain adaptation'' or ``generalization''.

\subsection*{Individual re-identification}

\justify Another important biometric is animal identity. The standard for identification of animal species and identity is DNA profiling~\cite{avise2012molecular}, which can be difficult to scale to large, distributed populations~\cite{kuhl2013animal,schneider2019past}. As an alternative to gene-based identification, manual tagging can be used to keep track of individual animals~\cite{kuhl2013animal,kays2015terrestrial}. Similar to counting and reconstruction (see next section), computer vision recently emerged as a powerful alternative for automatic individual identification~\cite{vidal2021perspectives,berger2017wildbook,parhamAnimalDetectionPipeline2018,schneider2019past}. The aim is to learn identity-bearing features from the appearance of animals. 
Identifying individuals from images is even more challenging than species recognition, since the distinctive body patterns of individuals might be subtle or not be sufficiently visible due to occlusion, motion blur, or overhead viewpoint in the case of aerial imagery. Yet, conventional~\cite{ Weideman_2020_WACV} and more recently DL-based~\cite{brust2017towards, schofield2019chimpanzee,schneider2019past} methods have reached strong performance for specific taxa, especially across small populations. 
Some species have individually-unique coat or skin markings that assist with re-identification: for example, accuracy exceeded 90\% in a study of 92 tigers across 8,000 video clips~\cite{li2020atrw}. However, effective re-identification is also possible in the absence of patterned markings: a study of a small group of twenty-three chimpanzees in Guinea applied facial recognition techniques to a multi-year dataset comprising 1 million images and achieved $>90$\% accuracy~\cite{schofield2019chimpanzee}. This study compared the DL model to manual re-identification by humans: where humans achieved identification accuracy between 20\% (novices) and 42 \% (experts) with an annotation time between one and two hours, the DL model achieved an identification accuracy of 84\% in a matter of seconds. The particular challenges for animal (re)-identification in wild populations are related to the difficulty of manual curation, larger populations, changes in appearance (e.g., due to scars, growth), few sightings per individual and the frequent addition of new individuals that may enter the system due to birth or immigration, therefore creating an ``open-set'' problem~\cite{bendale2016towards} wherein the model must deal with ``classes'' (individuals) unseen during training. The methods must have the ability to identify not only animals that have been seen just once or twice but also recognize new, previously unseen animals, as well as adjust decisions that have been made in the past, reconciling different views and biological stages of an animal.

\subsection*{Animal synthesis and reconstruction}

\justify 3D shape recovery and pose estimation of animals can provide valuable, non-invasive insights on wild species in their natural environment. The 3D shape of an individual can be related to its health, age or reproductive status; the 3D pose of the body can provide finer information with respect to posture attributes and allows, for instance, kinematic as well as behavioral analyses. For pose estimation, marker-less methods based on DL have tremendously improved over the last years and already impacted biology~\cite{Mathis2020DeepLT}. Various user-friendly toolboxes are available to extract the 2D posture of animals from videos (Fig.~\ref{fig:sentinel}d,e), while the user can define which body parts should be estimated (reviewed in~\cite{mathis2020primer}). Extracting a dense set of body surface points is also possible, as elegantly shown in~\cite{sanakoyeu2020transferring}, where the DensePose technique originally developed for humans was extended to chimpanzees. The reconstruction of the 3D shape and pose of animals from images often follows a model-based paradigm, where a 3D model of the animal is fit to visual data. Recent work defines the SMAL (Skinned Multi Animal Linear) model, a 3D articulated shape model for a set of quadruped families~\cite{zuffi20173d}. Biggs et al. built on this work for 3D shape and motion of dogs from video~\cite{biggs2018creatures} and for recovery of dog shape and pose across many different breeds~\cite{biggs2020left}. In~\cite{zuffi2019three} the SMAL model has been used in a DL approach to predict 3D shape and pose of the Grevy’s zebra from images. 3D shape models have been recently defined also for birds~\cite{wang2021birds}.
Image-based 3D pose and shape estimation methods provide rich information about individuals but require, in addition to accurate shape models, prior knowledge about the animal's 3D motion.

\vspace{-5pt}
\subsection*{Reconstructing the environment}

\justify Wildlife behavior and conservation cannot be dissociated from the environment animals evolve and live in. 
Studies have shown that animal observations like trajectories highly benefit from additional cues included in the environmental context~\cite{haalck2020towards}. Satellite remote sensing has become an integral part to study animal habitats, biological diversity and spatio-temporal changes of abiotic conditions~\cite{pettorelli2014satellite}, since it allows to map  quantities like land cover, soil moisture or temperature at scale. Reconstructing the 3D shape of the environment has also become central in behavior studies: for example, 3D reconstructions of kill sites for lions in South Africa revealed novel insights into the predator-prey relationships and their connection to ecosystem stability and functioning~\cite{davies2016effects}, while 3D spatial reconstructions shed light on the impact of forest structures on bat behavior~\cite{froidevaux2016field}.

\begin{figure*}
\includegraphics[width=.98\linewidth]{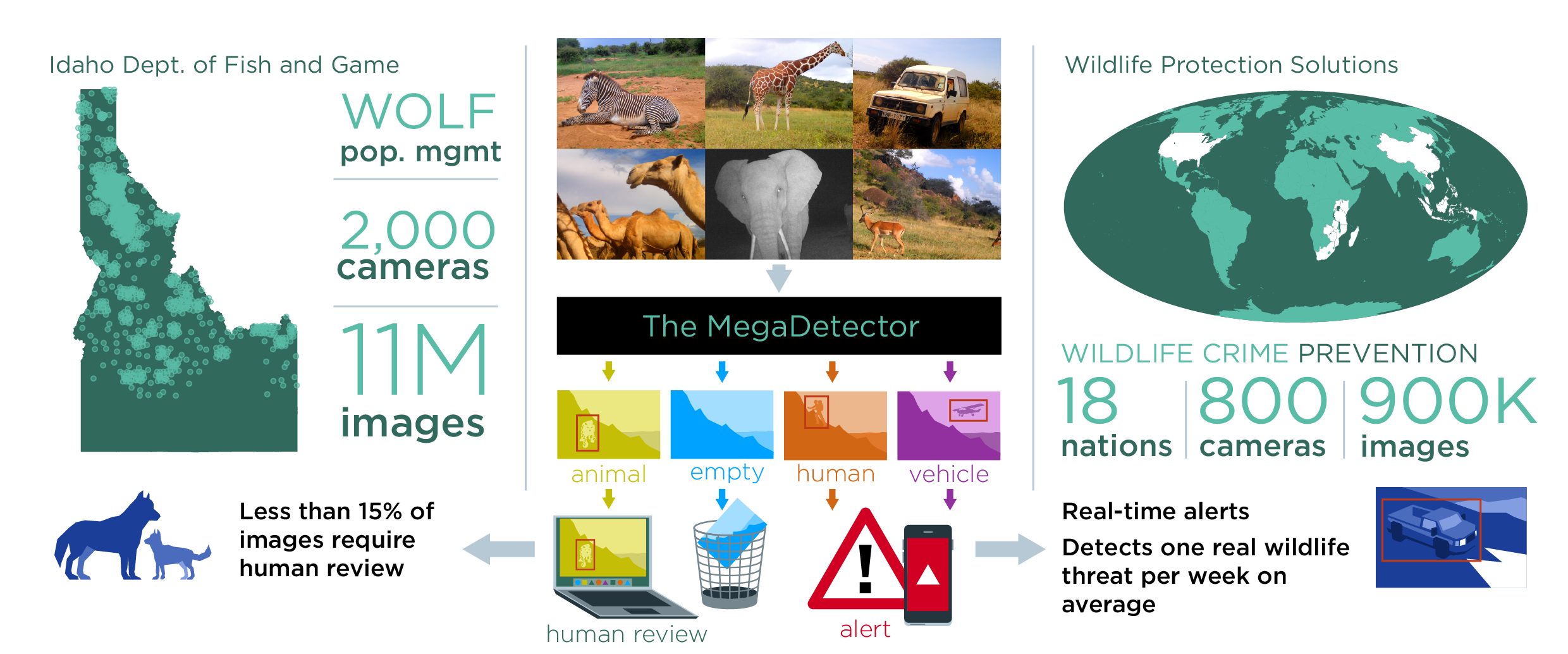}
\caption{{\textbf{AI for Wildlife Conservation in Practice: the MegaDetector}. The near-universal need of all camera trap projects to efficiently filter empty images and localize humans, animals, and vehicles in camera trap data, combined with the robustness to geographic, hardware, and species variability the MegaDetector provides due to its large, diverse training set makes it a useful, practical tool for many conservation applications out of the box. The work done by the Microsoft AI for Earth team to provide assistance running the model via hands-on engineering assistance, open source tools, and a public API have made the MegaDetector accessible to ecologists and a part of the ecological research workflow for over 30 organizations worldwide.}} 
\label{fig:MD}
\end{figure*}

\begin{pabox}[label={box2},nameref={Hardware}]{AI for Wildlife Conservation in Practice: the MegaDetector.}
One highly-successful example of open source AI for wildlife conservation is the \href{https://github.com/microsoft/CameraTraps/blob/master/megadetector.md}{Microsoft AI for Earth MegaDetector}~\cite{beery2019efficient} (Figure~\ref{fig:MD}). This generic, global-scale human, animal, and vehicle detection model works off-the-shelf for most camera trap data, and the publicly-hosted MegaDetector API has been integrated into the wildlife monitoring workflows of over 30 organizations worldwide, including the \href{https://www.wcs.org/}{Wildlife Conservation Society}, \href{https://sandiegozoo.org/sdzglobal/}{San Diego Zoo Global}, and \href{https://www.islandconservation.org/}{Island Conservation}. We would like to highlight two MegaDetector use cases, via Wildlife Protection Solutions (WPS) and the Idaho Department of Fish and Game (IDFG). WPS uses the MegaDetector API in real-time to detect threats to wildlife in the form of unauthorized humans or vehicles in protected areas. WPS connect camera traps to the cloud via cellular networks, upload photos, run them through the MegaDetector via the public API, and return real-time alerts to protected area managers. They have over 400 connected cameras deployed in 18 different countries, and that number is growing rapidly. WPS used the MegaDetector to analyze over 900K images last year alone, which comes out to 2.5K images per day. They help protected areas detect and respond to threats as they occur, and detect at least one real threat per week across their camera network. Idaho is required to maintain a stable population of protected wolves. IDFG relies heavily on camera traps to estimate and monitor this wolf population, and need to process the data collected each year before the start of the next season in order to make informed policy changes or conservation interventions. They collected 11 million camera trap images from their wolf cameras last year, and with the MegaDetector integrated into their data processing and analysis pipeline, they were able to fully automate the analysis of 9.5 million of those images, using model confidence to help direct human labeling effort to images containing animals of interest. Using the Megadetector halved their labeling costs, and allowed IDFG to label all data before the start of the next monitoring season, whereas manual labeling previously resulted in a lag of approximately five years from image collection to completion of labeling. The scale and speed of analysis required in both cases would not be possible without such an AI-based solution.
\label{box:md}
\end{pabox}

Such spatial reconstructions of the environment can either be extracted by using dedicated sensors such as LiDAR~\cite{risse2018software} or can be reconstructed from multiple images,  
either by stitching the images into a unified two-dimensional panorama (\emph{e.g.}, mosaicking~\cite{haalck2021embedded}) or by computing the three-dimensional environment from partially overlapping images (\emph{e.g.}, Structure from Motion~\cite{schonberger2016structure} or Simultaneous Localization and Mapping~\cite{mur2017orb}).
All these approaches have strongly benefited from recent ML advancements~\cite{kuppala2020overview}, but have seldom been applied for wildlife conservation purposes, where they could greatly help when dealing with images acquired by moving or swarms of sensors~\cite{lisein2013aerial}.
However, applying these techniques to natural wildlife imagery is not trivial. For example, unconstrained continuous video recordings at potentially high frame-rates will result in large image sets which require efficient image processing~\cite{haalck2021embedded}. Moreover, ambiguous environmental appearances and structural errors such as drift accumulate over time and therefore decrease the reconstruction quality~\cite{schonberger2016structure}. Last but not least, a variety of inappropriate camera motions or environmental geometries can result in so-called critical configurations which cannot be resolved by the existing optimization schemes~\cite{wu2014critical}. As a consequence, cues from additional external sensors are usually integrated to achieve satisfactory environmental reconstructions from video data~\cite{ferrer2007large}.

\subsection*{Modeling species diversity, richness and interactions}

\justify Analyses of biodiversity, represented by such measures as species abundance and richness, are foundational to much ecological research and many conservation initiatives. Spatially explicit linear regression models have been conventionally used to predict species and community distribution based on explanatory variables such as climate and topography~\cite{guisan2000predictive, lehmann_regression_nodate}. Non-parametric ML techniques like Random Forest~\cite{breiman2001random} have been successfully used to predict species richness and have shown significant error reduction with respect to the traditional counterparts used in ecology, for example in the estimation of richness distributions of fishes~\cite{parravicini2013global,smolinski2017spatial}, spiders ~\cite{vcandek2020machine}, and small mammals~\cite{baltensperger_predictive_2015}. Tree-based techniques have also been used to predict species interactions: for example, regression trees significantly outperformed classical generalized linear models in predicting plant-pollinator interactions~\cite{pichler2020machine}. Tree-based methods are well-suited to these tasks because they perform explicit feature ranking (and thus feature selection) and are able to model nonlinear relationships between covariates and species distribution. More recently, graph regression techniques were deployed to reconstruct species interaction networks in a community of European birds with promising results, including better causality estimates of the relations in the graph~\cite{faisal2010inferring}.

\section*{Attention points and opportunities}

Machine and deep learning are becoming necessary accelerators for wildlife research and conservation actions in natural reserves. We have discussed success stories of the application of approaches from ML into ecology and highlighted the major technical challenges ahead. In this section, we want to present a series of ``attention points" that highlight new opportunities between the two disciplines.

\subsection*{What can we focus on now?}

\justify State-of-the-art ML models are now being applied to many tasks in animal ecology and wildlife conservation. However, while an out-of-the-box application of existing open tools is tempting, there are a number of points and potential pitfalls that must be carefully considered to ensure responsible use of these approaches.

\begin{itemize}
    \item \emph{Inherent model biases and generalization.} Most ecological datasets suffer from some degree of geographic bias. For example, many open imagery repositories such as \href{https://artportalen.se/}{Artportalen.se}, \href{https://naturgucker.de/}{Naturgucker.de} and \href{https://waarneming.nl/}{Waarneming.nl} collect images from specific regions, and most contributions on iNaturalist~\cite{van2018inaturalist} (see Table~\ref{tab:csprojects}) come from the Northern hemisphere. Such biases need to be understood, acknowledged and communicated to avoid incorrect usage of methods or models that by design may only be accurate in a specific geographic region. Biases are not limited to the geographical provenance of images: the type of sensors used (RGB \emph{vs.} infrared or thermal), the species they depict and the imbalance in the number of individuals observed per species~\cite{beery2018recognition, van2018inaturalist} must also be considered when training or using models to avoid potentially catastrophic drop-offs in accuracy, and transparency around the training data and the intended model usage is a necessity~\cite{copas2019training}.

    \item \emph{Curating and publishing well-annotated benchmark datasets without doing harm.} The long-term advancement of the field will ultimately require the curation of large, diverse, accurately labeled, and publicly available datasets for ecological tasks with defined evaluation metrics and maintained code repositories. However, opening up existing datasets (and especially when using private-owned images acquired by non-professionals as in~\cite{parham2017animal}) is both a necessary and difficult challenge for the near future. Fostering a culture of individual and cross-institutional data sharing in ecology will allow ML approaches to improve in robustness and accuracy. Furthermore, proper credit has to be given to the data collectors, for example through appropriate data attribution and Digital Object Identifiers (DOIs) for datasets~\cite{copas2019training}. 

     \item \emph{Understanding the ethical risks involved}. Computer scientists must also be aware of the ethical and environmental risks of publishing certain types of datasets. Willingness to support `AI for good' actions is a noble and necessary action for computer scientists, but it is important to understand the limits of open data sharing in animal conservation in nature parks: in some cases it is imperative that the privacy of the data be preserved, for instance to avoid giving poachers access to locations of animals in near-real-time~\cite{lennox2020novel}. Security of rangers themselves is also at stake; for example the flight path of drones might be backtracked to reveal their location.

    \item \emph{Standards of quality control are urgently needed}. Accountability for open models needs to be better understood. The estimations of models remain approximations and need to be treated as such: population counts without uncertainty estimation can lead to erroneous and potentially devastating conclusions.  Increased quality control on the adequacy of a model to a new scientific question or study area is important and can be achieved by close cooperation between model developers (who have the ability to design, calibrate, and run the models at their best) and practitioners (who have the domain and local knowledge). Without such quality control measures, relying on model-based results is risky and could have difficult-to-evaluate impacts on research in animal ecology, as incorrect results hidden in a suboptimally trained model will become more and more difficult to detect. Computer scientists must be aware that errors by their models can lead to erroneous decisions on site that can be catastrophic for the population they are trying to preserve or for the populations that live at the border of human/wildlife conflicts.
 
     \item \emph{Environmental and financial costs of machine learning.} ML is not free. Training and running models with millions of parameters on large volumes of data requires powerful, somewhat specialized hardware. Purchasing prices of such machines alone are often prohibitively high especially for budget-constrained conservation organizations; programming, running and maintenance costs further add to the bill. Although cloud computing services exist that forgo the need of hardware management, they likewise pose per-resource costs that quickly scale to several thousands of dollars per month for a single virtual machine. Besides monetary costs, ML also uses significant amounts of energy: recently, it has been estimated that large, state-of-the-art models for understanding natural language emit as much carbon as several cars in their entire lifetime~\cite{strubell2019energy}. Even though the models currently used in animal ecology are far from such a carbon footprint, environmental costs of AI are often disregarded, as energy consumption of large calculations is still considered an endless resource (assuming that the money to pay for it is available). We believe this is a mistake, since disregarding environmental costs of ML models equals exchanging one source environmental harm (loss and biodiversity) for another (increase of emissions and energy consumption). Particular care needs to be paid to designing models that are not oversized and that can be trained efficiently. Smaller models are not only less expensive to train and use, their lighter computational costs allow them to be run on smaller devices, opening opportunities for real-time ML ``on the edge'' -- \emph{i.e.}, within the sensors themselves.

\end{itemize}

\subsection*{What's new: vast scientific opportunities lie ahead}

\justify In the previous sections, we describe the advances in research at the interface of ML, animal ecology and wildlife conservation. The maturity of the various detection, identification and recognition tools opens a series of interesting perspectives for genuinely novel approaches that could push the boundaries towards true integration of the disciplines involved.

\begin{itemize}
    \item \emph{Involving domain knowledge from the start.}
    The ML and DL fields have focused mainly on black box models that learn correlations from data directly, and domain knowledge has been repeatedly ignored in favor of generic approaches that could fit to any kind of dataset. Such universality of ML is now strongly questioned and the inductive bias of traditional DL models is challenged by new approaches that bridge domain knowledge, fundamental laws and data science. This ``hybrid models'' paradigm~\cite{Karpatne17,Reichstein2019} is one of the most exciting avenues in modern ML and promises real collaboration between domains of application and ML, especially when coupled with algorithmic designs that allow interpretation and understanding of the visual cues that are being used~\cite{samek2019explainable}. This line of interdisciplinary research is small but growing, with several studies published in recent years. A representative one is Context R-CNN~\cite{beery2020context} for animal detection and species classification, which leverages the prior knowledge that backgrounds in camera trap imagery exhibit little variation over time and that camera traps acquire data with low sampling frequency and occasional dropouts. By integrating image features over long time spans (up to a month), the model is able to increase mean species identification precision in the Snapshot Serengeti dataset~\cite{swanson2015snapshot} by 17.9\%. In another example~\cite{lutio2021digital}, the hierarchical structure of taxonomies, as well as locational priors, are leveraged to constrain plant species classification from iNaturalist in Switzerland, leading to improvements of state-of-the-art models of about 5\%. Similarly, \cite{mac2019presence} incorporate knowledge about the distribution of species as well as photographer biases into a DL model for species classification in images and report accuracy improvements of up to 12\% in iNaturalist over a baseline without such priors. Finally, \cite{gurumurthy_exploiting_2018} used expert knowledge of park rangers to augment sparse and noisy records of poaching activity, thereby improving predictions of poaching occurrence and enabling more efficient use of limited patrol resources in a Chinese nature reserve. These approaches challenge the dogma of ML models learning exclusively from data and achieve more efficient model learning (since base knowledge is available from the start and does not have to be re-learnt) and enhanced plausibility of the solutions (because the solution space can be constrained to a range of ecologically plausible outcomes).
 
    \item  \emph{Laboratories as development spaces.}
    In recent years modern ML has rapidly changed laboratory-based non-invasive observation of animals~\cite{Mathis2020DeepLT, mathis2020primer}. Neuroscience studies in particular have embraced novel tools for motion tracking, pose estimation (Figure~\ref{fig:sentinel}d, e) and behavioral classification (\emph{e.g.},~\cite{Datta2019ComputationalNA}). The high level of control (\emph{e.g.} of lighting conditions, sensor calibration, and environment) afforded by laboratory settings facilitated the rapid development of such tools, many of which are now being adopted for use in field studies of free-moving animals in complex natural environments~\cite{joska2021acinoset, graving2019deepposekit}. Additionally, algorithmic insights gained in the lab can be transferred back into the wild -- studies on short videos or camera traps can leverage lab-generated data that is arguably less diverse, but easier to control. This opens interesting research opportunities for the adaptation of lab-generated simulation to real world conditions, similar to what has been observed in the field of image synthesis for self driving~\cite{chen17photographic} and robotics~\cite{lee2020learning} in the last decade. Thus, laboratories rightly serve as the ultimate development space for such in-the-wild applications.
 
    \item \emph{Towards a new generation of biodiversity models}. Statistical models for species richness and diversity are routinely used to estimate abundances and study species co-occurrence and interactions. 
    Recently, DL methods have also started to be employed to model species' ecological niches~\cite{deneu2019evaluation,botella2021jointly}, facilitated by the development of machine-learning-ready datasets such as GeoLifeCLEF\footnote{\url{www.imageclef.org/GeoLifeCLEF2021}}.  GeoLifeCLEF curated a dataset of 1.9 million iNaturalist observations from North America and France depicting over 31,000 species, together with environmental predictors (land cover, altitude, climatic data, \emph{etc.}), and asked users to predict a ranked list of likely species per geospatial grid cell. The task is complex: only positive counts are provided, no absence data are available, and predictions are counted as correct if the ground truth species is among the 30 predicted with highest confidence. This challenging task remains an open challenge  -- the winners of the 2021 edition achieved only an approximate 26\% top-30 accuracy.
 
    A recent review of species distribution modeling aimed at ML practitioners \cite{beery2021species} provides an accessible entry point for those interested in tackling the challenges in this complex, exciting field at the intersection of ML, remote sensing data analysis and statistical modeling. Open challenges in the field include increasing the scale of joint models geospatially, temporally, and taxonomically, building methods that can leverage multiple data types despite bias from non-uniform sampling strategies, incorporating ecological knowledge such as species dispersal and community composition, and expanding methods for the evaluation of these models.

 \end{itemize}
 
    Finally, we wish to re-emphasize that the vision described here cannot be achieved without interdisciplinary thinking: for all these exciting opportunities, processing big ecological data is necessitating analytical techniques of such complexity that no single ecologist can be expected to have all the technical expertise (plus domain knowledge) required to carry out groundbreaking studies~\cite{williams2020optimizing}. Cross-disciplinary collaborations are undeniably a critical component of ecological and conservation research in the modern era. Mutual understanding of the field-specific vocabularies, of the fields' expectations and of the implications and consequences of research ethics are within reach, but require open dialogues between communities, as well as cross-domain training of new generations.

\section*{Conclusions}
Animal ecology and wildlife conservation need to make sense of large and ever-increasing streams of data to provide accurate estimations of populations, understand animal behavior and fight against poaching and loss of biodiversity. Machine and deep learning (ML; DL) bring the promise of being the right tools to scale local studies to a global understanding of the animal world. \\

In this \textit{Perspective}, we presented a series of success stories at the interface of ML and animal ecology. We highlighted a number of performance improvements that were observed when adopting solutions based on ML and  new generation sensors. Although often spectacular, such improvements require ever-closer cooperation between ecologists and ML specialists, since recent approaches are more complex than ever and require strict quality control and detailed design knowledge. We observe that skillful applications of state-of-the-art ML concepts for animal ecology now exist, thanks to corporate (\emph{e.g.}, Wildlife Insights) and research (AIDE, MegaDetector, DeepLabCut) efforts, but that there is still much room (and need) for genuinely new concepts pushed by interdisciplinary research, in particular towards hybrid models and new habitat distribution models at scale.

Inspired by these observations, we provided our perspective on the missing links between animal ecology and ML via a series of attention points, recommendations and vision on future exciting research avenues. We strongly incite the two communities to work hand-in-hand to find digital, scalable solutions that will elucidate the loss of biodiversity and its drivers and lead to global actions to preserve nature. Computer scientists have yet to integrate ecological knowledge such as underlying biological processes into ML models, and the lack of transparency of current DL models has so far been a major obstacle to incorporating ML into ecological research. However, an interdisciplinary community of computer scientists and ecologists is emerging, which we hope will tackle this technological and societal challenge together.

\section*{References}

 \bibliography{wild_refs,tanya}
 
\balance
\section*{Acknowledgments}
 
We thank Mike Costelloe for assistance with figure design and execution. 
SB would like to thank the Microsoft AI for Earth initiative, the Idaho Department of Fish and Game, and Wildlife Protection Solutions for insightful discussions and providing data for figures.

\textbf{Funding:}
MCC and TBW were supported by the National Science Foundation (IIS 1514174 \& IOS 1250895). MCC received additional support from a Packard Foundation Fellowship (2016-65130), and the Alexander von Humboldt Foundation in the framework of the Alexander von Humboldt Professorship endowed by the Federal Ministry of Education and Research.
CVS and TBW were supported by the US National Science Foundation (Awards 1453555 and 1550853).
SB was supported by the National Science Foundation Grant No. 1745301 and the Caltech Resnick Sustainability Institute. 
IDC acknowledges support from the ONR (N00014- 19-1-2556), and IDC, BRC, MW and MCC from, the Deutsche Forschungsgemeinschaft (German Research Foundation) under Germany’s Excellence Strategy–EXC 2117-422037984. MWM is the Bertarelli Foundation Chair of Integrative Neuroscience.

\textbf{Reprints:}
Fig.~2 part B was reprinted in part from \cite{de2021timely}. Fig.~2 part C was reprinted in part from~\cite{walter_trex_2021}. Fig.~2 part D was reprinted from~\cite{graving2019deepposekit}. Fig.~2 part E was reprinted from~\cite{joska2021acinoset}. All reprinted images are permitted for this use due to original publication under the Creative Commons Attribution License (CC-BY 4.0). Satellite image in Fig.~3 is a Sentinel-2 (ESA) image courtesy of the U.S. Geological Survey.

\textbf{Notice:}
Any opinions, findings, and conclusions or recommendations expressed in this material are those of the author(s) and do not necessarily reflect the views of the funding agencies. The authors declare no competing interests.

\section*{Author contributions}
 
DT coordinated the writing team; DT, BK, SB and BC structured and organized the paper with equal contributions; all authors wrote the text; BC created the figures.

\begin{table*}[!h]
\centering
\small{
\caption{Resources for machine and deep learning-based wildlife conservation}
\label{tab:tools}
\begin{tabular}{p{1.9cm}|p{11cm}|c}
\hline
Name& Description & URL \\\hline
AIDE~\cite{kellenberger2020aide} & 
\textbf{Tasks: Annotation; detection; classification; segmentation}\newline
 Free, open source, web-based, collaborative labeling platform specifically designed for large-scale ecological image analyses. Users can concurrently annotate up to billions of images with labels, points, bounding boxes, or pixel-wise segmentation masks. AIDE tightly integrates ML models through Active Learning~\cite{settles2012active}, where annotators are asked to provide inputs where the model is the least confident. AIDE further offers functionality to share and exchange trained ML models with other users of the system for collaborative annotation efforts in image campaigns across the globe. & \href{https://github.com/microsoft/aerial_wildlife_detection}{GitHub} \\\hline

MegaDetec-
tor~\cite{beery2019efficient}&
\textbf{Tasks: Detection}\newline
Free and open source detector based on deep learning hosted by Microsoft AI4Earth. The current model is 
trained with the TensorFlow Object Detection API using several hundred thousand camera trap images labeled with bounding boxes from a variety of ecosystems. The model identifies animals (not species-specific), humans, and vehicles, and is robust to novel sensor deployment locations and taxa not seen during training. Updates of the model, trained with additional data, are periodically released. Microsoft AI4Earth provides support to assist ecologists in using the model, including a public API for batch inference, and integration with commonly-used camera trap data management platforms such as TimeLapse and Camelot.
& \href{https://github.com/microsoft/CameraTraps/blob/master/megadetector.md}{GitHub} \\\hline
Wildbook \cite{berger2017wildbook} & \textbf{Tasks: Individual Re-Identification}\newline
Wildbook blends structured wildlife research with artificial intelligence, community science, and computer vision to speed population analysis and develop new insights to help fight extinction. They host community-run individual re-identification systems and global data repositories for a broad and expanding set of species, including Grevy's Zebra, Whale sharks, Manta Rays, and many more.& \href{https://www.wildme.org/#/wildbook}{URL} 
\\\hline
Wildlife Insights~\cite{ahumada2020wildlife} & 
\textbf{Tasks: Filtering} \newline
Large-scale platform for camera trap data management with computer vision in the backend. Currently open for whitelisted users, extensible via a waitlist. Wildlife Insights filters blank images and provides species identification for images that the computer vision model scores highly, allowing expert ecologists to focus on labeling only challenging images.& \href{https://www.wildlifeinsights.org/home}{URL} \\
\hline
DeepLabCut
\cite{mathis2018deeplabcut} & \textbf{Tasks: Pose estimation and behavioral analysis.}\newline Free and open source pose estimation toolbox based on deep learning. Pre-trained models (for instance for primate faces and bodies, as well as quadruped) as well as a light-weight, real-time version are available.  & \href{https://github.com/DeepLabCut/DeepLabCut}{GitHub} \\
\hline
DeepPoseKit
\cite{graving2019deepposekit} & \textbf{Tasks: Pose estimation and behavioral analysis.}\newline Free and open source pose estimation toolbox based on deep learning. & \href{https://github.com/jgraving/DeepPoseKit}{GitHub} \\
\hline
\end{tabular}
}
\end{table*}

\begin{table*}[!h]
\centering
\caption{Examples of community science projects in digital wildlife conservation
}
\label{tab:csprojects}
\begin{tabular}{p{4cm}|p{3cm}|p{3cm}|p{2.5cm}|c}
\hline
Name&  Spatial coverage & Sensor & Task & Ref. \\\hline
\href{https://www.inaturalist.org}{iNaturalist} & Global & Human photographers & Classification Detection& \cite{van2018inaturalist} \\\hline
\href{http://www.epfl.ch/savmap}{SAVMAP} & Kuzikus reserve,  Namibia & UAV images & Detection   & \cite{ofli2016combining} \\\hline
\href{https://www.zooniverse.org/projects/zooniverse}{Zooniverse} &Global&Images, Text, Video& Classification
Detection& \cite{simpson2014zooniverse} \\\hline
\href{https://www.brc.ac.uk/irecord/}{iRecord} &United Kingdom& Photographic records & Classification& \cite{pocock2015biological} \\\hline
\href{http://www.greatgrevysrally.com}{Great Grevy's Rally}  &Northern Kenya&Safari pictures& Classification Detection Identification& \cite{parham2017animal}\\\hline
\end{tabular}
\end{table*}

\begin{table*}[!h]
    \scriptsize{
        \centering
        \caption{Most common ML models}
        {\begin{tabular}{p{2cm}|p{3.5cm}|p{2cm}|p{3.5cm}|p{4cm}}
             Model & Description & Output & Advantages & Limitations  \\\hline
                 \multicolumn{5}{l}{\textit{Traditional machine learning models}}\\\hline
             Bayesian estimation & Maximum a posteriori estimation of predictions; data are assumed to be drawn from an a priori known (``prior'') distribution & Classification, regression & Can include prior knowledge about data distribution & Hyperparameter tuning can be expensive, performance depends on quality of features\\
             Decision tree & Iterative binary split of data points according to input variables or features & Classification, regression & Very simple, intuitive and interpretable model, split thresholds can be learned from data or manually defined & Highly prone to overfitting under too many splits (large tree depth); weak performance and poor generalization capabilities if single tree (see Random Forest below); does not provide probability measures\\
             Random Forest~\cite{breiman2001random} & Ensemble of decision trees, with each tree receiving a randomized subset of data points and variables to operate on & Classification, regression & Requires little training data, can model non-linear relationships by design & Limited scalability, performance depends on quality of features \\
             Support Vector Machine~\cite{cortes1995support} & Binary classifier based on maximum margin theory & Classification, regression & Requires very little training data & Binary predictions only in original formulation; can only model non-linear relationships through kernels; performance depends on quality of features \\

         \hline
             \multicolumn{5}{l}{\textit{Deep learning models}}\\\hline
             Artificial Neural Network (ANN) & Model that applies a sequence of layers, each composed of neurons that receive all values of a data point (first layer) or outputs of the previous layer and calculate a weighted and biased combination as an output. & Classification, regression & Universal approximator, can reproduce very nonlinear behavior & Poor scalability to large data points like images; overfitting and need for early stopping in training \\
             Convolutional Neural Network (CNN~\cite{LeCun15}) & Form of ANN with convolution operators and generally large number of layers & Arbitrary (classification, regression, segmentation, mixtures, \emph{etc.}) & Excellent performance in most machine learning tasks; high versatility & Computationally expensive; generally requires large amounts of training data \\
             Vision Transformers~\cite{dosovitskiy2020image} & Most recent alternative to CNNs that replaces convolutional layers with spatial attention modules & Arbitrary & Extremely high performance in some tasks & Extremely high computational requirements; recent method with research still ongoing \\
             Recurrent Neural Network (RNN) & Form of ANN that ingests time series data in a point-wise manner, with each output (intermediate or final) depending on the current input as well as the previous output. RNNs can also be convolutional. & Arbitrary, on time series & Excellent performance in most machine learning tasks; high versatility & Computationally expensive; generally requires large amounts of training data; signals at early time steps in long time series may get lost in plain RNNs \\
             Long Short-Term Memory (LSTM~\cite{hochreiter1997long}) and Gated Recurrent Unit (GRU~\cite{cho2014learning}) & Form of RNN with dedicated ``gates'' that learn to memorize relevant signals in a time series & Arbitrary, on time series & Excellent performance in particular for long time series data & Computationally expensive; generally requires large amounts of training data \\
             \hline
        \end{tabular}}
    }
    \label{tab:mlModels}
\end{table*}

\onecolumn

\section*{Glossary}
\textbf{Glossary on the most important Machine Learning (ML) terms used in this article}

\begin{center}
\begin{scriptsize}
\begin{longtable}{p{3.6cm}|p{13cm}}
ML term & Definition\\\hline

Artificial Intelligence (AI) & The concept of a machine being able to perform higher-level, semantic reasoning.\\
Big data & Many definitions exist~\cite{de2016formal}, but we cast ``big data'' as \emph{information content for analyses whose volumes are too large to handle for users with conventional hardware}. Many sensors addressed produce ``big data'', in particular remote sensing, social media and camera trap networks. Analysis of such volumes of data quickly becomes intractable for conventional ML methods, in particular if the study area of interest exceeds regional ecosystems.\\
Classification & Assigning an entire image or video to a single category.\\
Computer Vision & Performing image manipulation and understanding tasks with a machine, oftentimes involving ML.\\
Convolutional Neural Network (CNN) & Deep learning models that contain at least one convolution layer. In such layers, neurons are organized into banks of filters that are convolved with the inputs (\emph{i.e.}, the same filter weights are applied across multiple locations in the image). This allows reducing the number of required neurons while also providing a limited amount of translation invariance.\\
Data science & Like ``big data'', ``data science'' is a less-well-defined term, denoted here as an inter- or multidisciplinary research field on automated information extraction from observations or other content sources.\\
Deep learning & Family of prediction models that consist of neurons, grouped into three or more sequential layers, where each neuron receives the output from one (or more) previous neurons and itself predicts an output, consisting of weighted combinations of its inputs.\\
(visual) Descriptor & Higher-level statistics extracted from data that are supposed to summarize, or pronounce, more abstract differences within the data point to facilitate the task of the subsequent ML model, also called ``feature''. For example, a common descriptor used in traditional vegetation mapping on remote sensing imagery is the Normalized Difference Vegetation Index (NDVI), whose values are highly contrastive between vegetated and non-vegetated areas than bare pixel values alone. Traditional ML algorithms require manual definition and calculation of such features, whereas deep learning methods learn them automatically in the training process.\\
Detection & localizing the area within an image that corresponds to a category of interest, usually represented by a rectangular ``bounding box'' -- the tightest box that could be drawn around that object while still containing all of its pixels.\\
Domain Adaptation & Methods to describe, evaluate, and/or tackle the challenge of out-of-domain data.\\
Detection rate & See ``recall''.\\
False positive & Incorrect prediction of a data point, object, or background area (\emph{e.g.} in an image) as a certain class.\\
Feature & See ``(visual) Descriptor''.\\
Fine-grained classification & Label classes are denoted as ``fine-grained'' if they belong to a common supercategory (\emph{e.g.}, ``American Robin'' and ``Guineafowl'' both belong to the supercategory ``bird''). Fine-grained classification can be challenging if  categories exhibit similar visual properties.\\
Individual identification & Recognizing unique instances of an object in an image or video (frame). Individual identification is usually performed through recognizing of unique visual cues that serve as ``fingerprints'' for an individual, such as the striping pattern of zebra or dot pattern on the back of whale shark individuals.\\
Inference & The act of performing prediction with a (trained) Machine ML model.\\
Instance Segmentation & Grouping every pixel in an image with the other pixels corresponding to that same \emph{instance} or object. If the image contained seven lions, each lion would be categorized with a different pixel label, even if the lions' pixel masks touch each other.\\
Localization & Identifying the position of an object within an image or video (frame). Unlike Detection, localization may not always include estimation of the full extents of an object, \emph{e.g.} through a bounding box, but might be limited to spatial coordinates of the object's center.\\
Loss function & Numerical criterion that measures the disagreement between an ML model prediction and the Ground Truth labels. For example, the \emph{cross-entropy loss function} returns the negative log likelihood between a predicted model probability and the label class.\\
Machine Learning (ML) & The ability of a computer to perform prediction tasks by learning from data (\emph{i.e.}, without primarily relying on hard-coded cascades of rules).\\
Semantic Segmentation & Assigning every pixel in an image to a specific class, \emph{i.e.}, all ``lion pixels'' would be labeled as such, regardless of the actual individual they belong to.\\
Semi-supervised learning & Training an ML model on data for which only a small subset contains labels.\\
Supervised learning & Training an ML model on data that consists of inputs (\emph{e.g.}, images) and labels (\emph{e.g.}, species names, bounding boxes).\\
Object detection & See ``Detection''.\\
Open-set & Scenario where a dataset may exhibit categories at test time that were unseen during ML model training. For example, a model for individual identification may be presented with images of an individual that got newly introduced to the area after training, and needs to be able to recognize it as a new individual accordingly.\\
Out-of-domain & Data that is not drawn from the identical set that an ML model was trained on. A good example of this would be images from a camera trap that was not seen during training.\\
Overfitting & Training an ML model to achieve (near-) perfect accuracy on the training set, but unacceptable accuracy on the validation or test set. Overfitting can occur if the model has too many free parameters or if the training set is not representative enough. See also ``Underfitting''.\\
Pose Estimation & 2D: predicting the pixel location of known parts of an object, for example, localizing the nose, eyes, joints, and tail of a lion. 3D: predicting the parts location in space, or predicting the 3D rotation of an articulated animal skeleton.\\
Posture Estimation & See ``Pose Estimation''.\\
Precision & Class-wise measure of exactness of ML model predictions. A precision of 1.0 means that every prediction made by a model is correct, while one approaching 0.0 means that there is a high number of wrong predictions (see ``false positive'').\\
Recall & Class-wise measure of completeness of ML model predictions. A recall of 1.0 means that every data point with a given true label class has been correctly predicted as such by the model, while a recall of 0.0 means that the model has missed all data points of that class.\\
Tracking & Localizing individual objects and correctly match them between frames throughout a video or temporal sequence of images.\\
Training & Altering the free (learnable) parameters of an ML model to optimize it to the training dataset, usually performed by minimizing values of a Loss function.\\
Underfitting & An ML model underfits the training set if it cannot appropriately capture the data distribution, resulting in unacceptable accuracy. Underfitting usually occurs if the model does not have a sufficient number of free parameters. See also ``Overfitting''.\\
Unsupervised learning & Training an ML model on data that only consists of inputs, but not of labels.\\
\hline
\label{tab:glossary}
\end{longtable}
\end{scriptsize}
\end{center}

\end{document}